\documentclass[conference]{IEEEtran}
\IEEEoverridecommandlockouts
\usepackage{cite}
\usepackage{amsmath,amssymb,amsfonts}
\usepackage{algorithmic}
\usepackage{graphicx}
\usepackage{textcomp}
\usepackage{float}
\usepackage{stfloats}
\usepackage{xcolor}
\usepackage{lipsum,mwe,cuted}
\usepackage{caption}
\usepackage{todonotes}
\newcommand{\etal}{\textit{et al}.}
\def\BibTeX{{\rm B\kern-.05em{\sc i\kern-.025em b}\kern-.08em
    T\kern-.1667em\lower.7ex\hbox{E}\kern-.125emX}}

\makeatother

\begin{document}

\title{Faces à la Carte: Text-to-Face Generation via Attribute Disentanglement\\
}

\author{\IEEEauthorblockN{1\textsuperscript{st}Tianren Wang}
\IEEEauthorblockA{\textit{ITEE} \\
\textit{The University of Queensland}\\
Brisbane, Australia \\
tianren.wang@uqconnect.edu.au}
\and
\IEEEauthorblockN{2\textsuperscript{nd}Teng Zhang}
\IEEEauthorblockA{\textit{ITEE} \\
\textit{The University of Queensland}\\
Brisbane, Australia\\
}
\and
\IEEEauthorblockN{3\textsuperscript{rd}Brian C. Lovell}
\IEEEauthorblockA{\textit{ITEE} \\
\textit{The University of Queensland}\\
Brisbane, Australia\\
}
}

\maketitle

\begin{strip}
  \centering{\includegraphics[width=\textwidth]{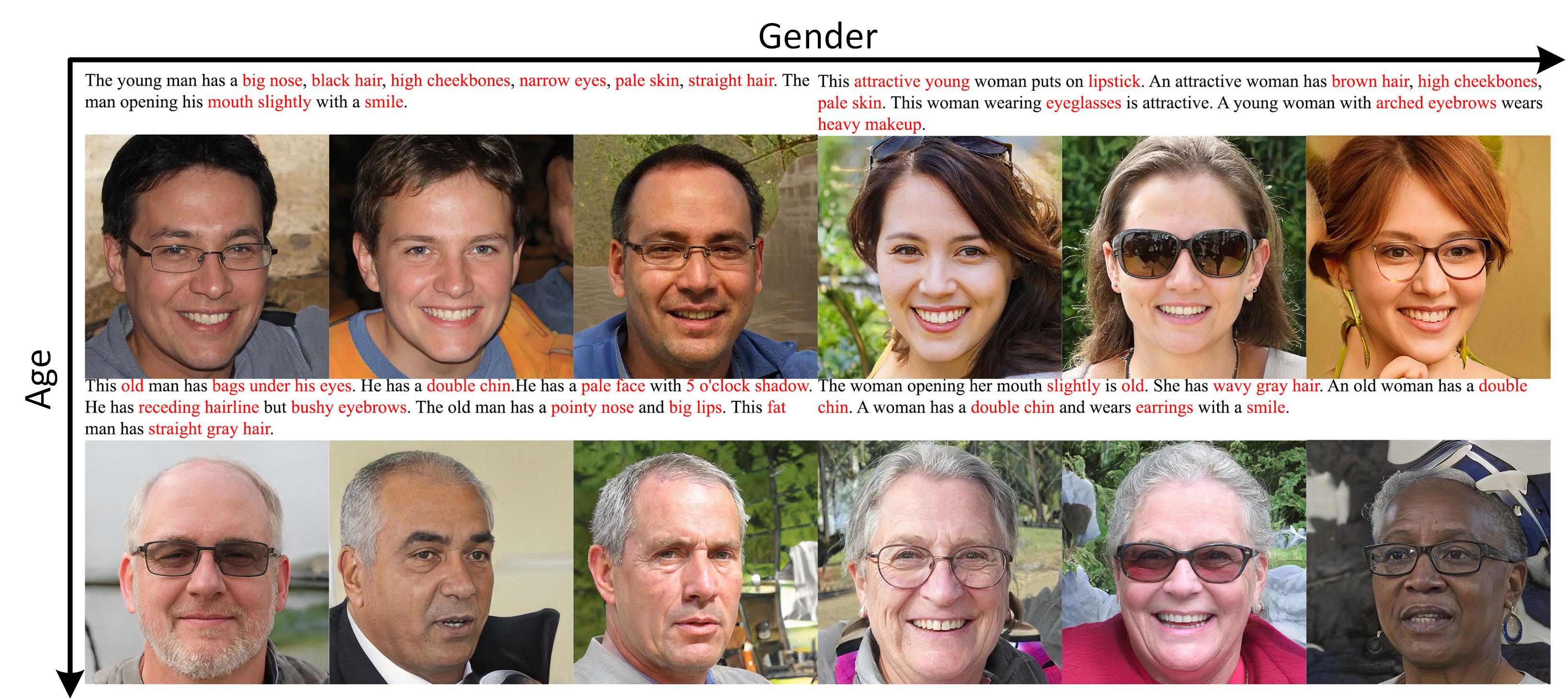}}
  \captionof{figure}{Several examples of synthesised images of our model.
    We select four groups of images which are arranged with respect to gender and age.
    The highlighted features in the textual descriptions are all rendered in the images.
    The images also exhibit diversity in terms of the other unspecified features.
    }
  \label{fig1}
  \vspace{+0.4 cm}
\end{strip}

\begin{abstract}
  Text-to-Face (TTF) synthesis is a challenging task with great potential for diverse computer vision applications.
  Compared to Text-to-Image (TTI) synthesis tasks, the textual description of faces can be much more complicated and detailed due to the variety of facial attributes and the parsing of high dimensional abstract natural language.
  In this paper, we propose a Text-to-Face model that not only produces images in high resolution (1024$\times$1024) with text-to-image consistency, but also outputs multiple diverse faces to cover a wide range of unspecified facial features in a natural way.
  By fine-tuning the multi-label classifier and image encoder, our model obtains the vectors and image embeddings which are used to transform the input noise vector sampled from the normal distribution.
  Afterwards, the transformed noise vector is fed into a pre-trained high-resolution image generator to produce a set of faces with the desired facial attributes.
  We refer to our model as TTF-HD.
  Experimental results show that TTF-HD generates high-quality faces with state-of-the-art performance.

\end{abstract}


\section{Introduction}
With the advent of Generative Adversarial Networks (GAN) \cite{b1}, image generation has made huge strides in terms of both image quality and diversity.
However, the original GAN model \cite{b1} cannot generate images tailored to meet design specifications.
To this end, many conditional GAN models have been proposed to fit different task scenarios \cite{b2,b3,b4,b5,b6,b7,b8}.
Among these works, Text-to-Image (TTI) synthesis is a challenging yet less studied topic.
TTI refers to generating a photo-realistic image which matches a given text description.
As an inverse image captioning task, TTI aims to establish an interpretable mapping between image space and the text semantic space.
TTI has huge potential  and can be used in many applications including photo editing and computer-aided design.
However, natural language is high dimensional information which is often less specific but also much more abstract than images.
Therefore, this research problem is quite challenging.

Just like TTI synthesis, the sub-topic of Text-to-Face (TTF) synthesis also has practical value in areas such as crime investigation and also biometric research.
For example, the police often need professional artists to sketch pictures of suspects based on the descriptions of the eyewitnesses.
This task is time-consuming, requires great skill and often results in inferior images.
Many police may not have access to such professionals.
However, with a well-trained Text-to-Face model, we could quickly produce a wide diversity of high-quality photo-realistic pictures based simply on the descriptions of eyewitnesses.
Moreover, TTF can be used to address the emerging issues of data scarcity arising from the growing ethical concerns regarding informed consent for the use of faces in biometrics research.


A major challenge of the TTF task is that the linkage between face images and their text descriptions are much looser than for, say, the bird and flower images commonly used in TTI research.
A few sentences of description are hardly adequate to cover all the variations of human facial features.
Also, for the same face image, different people may use quite different descriptions.
This increases the challenge of finding mappings between these descriptions and the facial features.
Therefore, in addition to the aforementioned two criteria, a TTF model should have the ability to produce a group of images with high diversity conditioned on the same text description.
In a real-world application, a witness could choose one image among several possible images which they believe is the closest the suspect.
Diversity is also very important for biometric researchers to obtain enough photos of rare ethnicities and demographics when synthesising ethical face datasets that do not require informed consent.

Note that there are raising ethical concerns in the face research community regarding some questionable behaviours such as harvesting faces from internet without user consent.
Constructing face datasets with racial bias where minorities are often neglected, and offensive slurs from the meta-data of faces harvested from the internet.
As a pioneer work targeting on resolving these ethical issues, our research start from resolving the issue of violating user consent first and, to some extent, combating the racial bias problem by providing ability to generated desired faces conditioned by text description.

Therefore, to meet these demands, we proposed a model which includes a novel TTF framework satisfying: 1) high image quality; 2) improved consistency of synthesised images and their descriptions; and 3) the ability to generate a group of diverse faces from the same text description.

To achieve these goals, we propose a pre-trained BERT \cite{b9} multi-label model for natural language processing.
This model outputs sparse text embeddings of length 40.
We then fine-tune a pre-trained MobileNets \cite{b10} model using CelebA's \cite{b11} training data where the images have paired labels.
Next, we predict labels from the input images.
Then we structure a feature space with 40 orthogonal axes based on the noise vectors and the predicted labels.
After this operation, the input noise vectors can be moved along specified directions to render output images which exhibit the desired features.
Last, but certainly not least, we use the state-of-the-art image generator, StyleGAN2 \cite{b12}, which maps the noise vectors into a feature disentangled latent space, to generate high-resolution images.
As Fig.
\ref{fig1} shows, the synthesised images match the features of the description while exhibiting both good diversity and excellent image quality.

\section{Contributions}
Note that this work is one part of the EDITH project which has been reviewed by the Office of Research Ethics and is deemed to be outside the scope of ethics review under the National Statement on Ethical Conduct in Human Research and University policy.

Our work has the following main contributions.
\begin{itemize}
  \item Proposes a novel TTF-HD framework comprising a multi-label text classifier, an image label encoder, and a feature-disentangled image generator to generate high-quality faces with a wide range of diversity.
  \item Adds a novel 40-label orthogonal coordinate system to guide the trajectory of the input noise vector.
  \item Uses state-of-the-art StyleGAN2 \cite{b12} as the generator to map the manipulated noise vectors into the disentangled feature space to generate 1024$\times$1024 high-resolution images.
\end{itemize}

This paper is continued as follow.
In Section 2, we review the important works in TTI, TTF, and models of the generators.
In Section 3, we describe our proposed framework in detail.
In Section 4, experimental results are presented both qualitatively and quantitatively and an ablation study is conducted to show the importance of the vector manipulating operations.
In Section 5, we conclude our work by summarising our contributions and the limitations of our approach.

\begin{figure*}[t!]
  \centerline{\includegraphics [width=\textwidth]{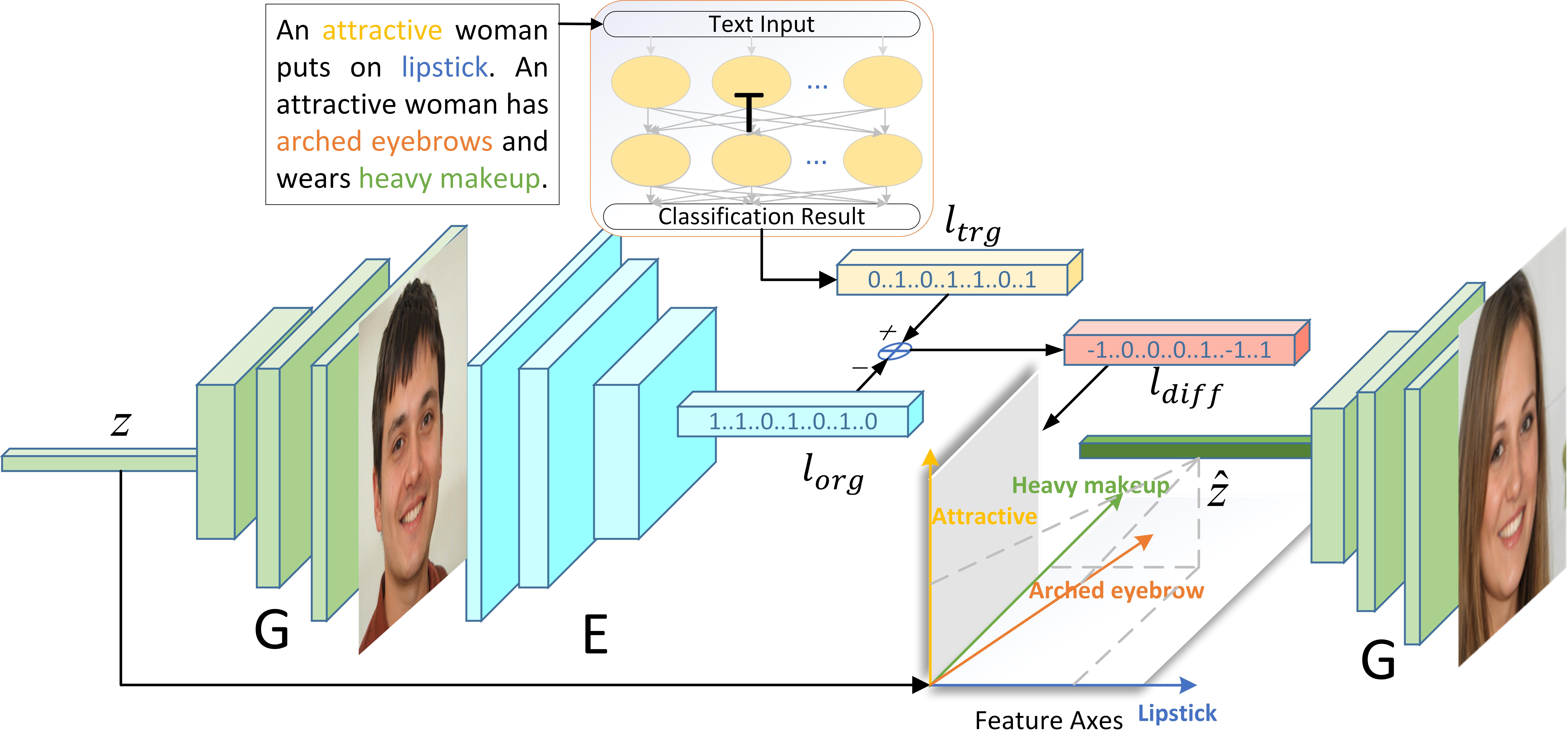}}
  \caption{TTF-HD diagram.
    The text is fed into the multi-label classifier $\boldsymbol{T}$ which then outputs a text vector $l_{trg}$ that represents 40 facial attributes.
    The image generator $\boldsymbol{G}$ firstly synthesises an image from a random noise vector $\boldsymbol{z}$.
    Then the image encoder $\boldsymbol{E}$ outputs the image embeddings $l_{org}$.
    The differentiated embedding $l_{diff}$ is used to manipulate the original noise vector from $\boldsymbol{z}$ to $\hat{\boldsymbol{z}}$.
    Finally, the generator synthesises an image with the desired features from $\hat{\boldsymbol{z}}$.
  }
  \label{fig2}
\end{figure*}

\section{Related Works}

\subsection{Text-to-Image Synthesis}

In the area of TTI, Reel \etal\
\cite{b6} first proposed to take advantage of GAN, which includes a text encoder and an image generator and they simply concatenated the text embedding to the noise vector as input.
Unfortunately, this model failed to establish good mappings between the keywords and the corresponding image features.
Moreover, due to the final results being directly generated from the concatenated vectors, the image quality was so poor that images were be easily spotted as fake.
To address these two issues, StackGAN \cite{b7} proposed to generate images hierarchically by utilising two pairs of generators and discriminators.
Later, Xu \etal\
proposed AttnGAN \cite{b8}.
By introducing the attention mechanism, this model successfully matched keywords with the corresponding image features.
Their interpolation experimental results indicated that the model could correctly render the image features according to the selected keywords.
This model works remarkably well in translating bird and flower descriptions.
However, in these cases the descriptions are mostly just one sentence.
If the descriptions are longer, the efficacy of text encoding deteriorates because the attention map becomes harder to train.

\subsection{Text-to-Face Synthesis}
Compared to the number of works in TTI, the published works in TTF are far fewer.
The main reason is that a face description has a much weaker connection to facial features compared to that of, say, bird or flower images.
Typically, the descriptions of birds and flowers are primarily about the colour of the feathers or petals.
Descriptions of faces can be much more complicated with gender, age, ethnicity, pose, and other important facial attributes.
Moreover, most of the TTI models are trained on Oxford-102 \cite{b13}, CUB \cite{b14}, and COCO \cite{b15} which are not face image datasets.
When dealing with faces, the only face dataset that is suitable is Face2text \cite{b16} which has only five thousand pairs of samples --- this not large enough to train a satisfactory model.

With all of the challenges mentioned above, there are still several inspiring works engaging in Text-to-Face synthesis.
In a project named T2F \cite{b17}, Akanimax proposed to encode the text descriptions into a summary vector using the LSTM network.
ProGAN \cite{b18} was adopted as the generator of the model.
Unfortunately, the final output images exhibited poor image quality.
Later, the author improved his work, which he named T2F 2.0, by replacing ProGAN with MSG-GAN \cite{b19}.
As a result, both image quality and image-text consistency improved considerably, but the output showed low diversity with regard to facial appearance.

To address the data scarcity issue, O.R.
Nasir \etal\ \cite{b20} proposed to utilise the labels of CelebA \cite{b11} to produce structured pseudo-text descriptions automatically.
In this way, the samples in the dataset are paired with sentences which contain positive feature names separated by conjunctions and punctuation.
The results are 64$\times$64 pixel images showing a certain degree of diversity in appearance.
The best output image quality so far is from Chen \etal\ \cite{b23} which also adopted the model structure of AttnGAN \cite{b8}.
Therefore, this work has the same issues with text encoding mentioned previously.

\subsection{Feature-Disentangled Latent Space}
Conventionally, the generator will produce random images from noise vectors sampled from a normal distribution.
However, we desire to control the rendering of the images in response to the feature labels.
To do this, Chen \etal\ \cite{b24} proposed to disentangle the desired features, by maximising the mutual information between the latent code $\boldsymbol{c}$ of the desired features and the noise vector $\boldsymbol{x}$.
In his experiments, he introduced a variation distribution $Q(\boldsymbol{c} \vert \boldsymbol{x})$ to approach $P(\boldsymbol{c}\vert \boldsymbol{x})$.
Finally, the latent code indicates that it has managed to learn interpretable information by changing the value in a certain dimension.
However, the latent code in this work has only 3 or 4 dimension; we require 40 features, which is much more complicated.
Later, Karras \etal\ \cite{b21} established a novel style-based generator architecture, named StyleGAN, which does not take the noise vector as input like the previous works.
The input vector is mapped into an intermediate latent space through a non-linear network before being fed into the generator network.
The non-linear network consists of eight fully connected layers.
A benefit for such a setting is that the latent space does not have to support sampling according to any fixed distribution \cite{b21}.
In other words, we have more freedom to combine desired features.

\section{Proposed Method}
Our proposed model, named TTF-HD, comprises a multi-label classifier $\boldsymbol{T}$, image encoder $\boldsymbol{E}$, and a generator $\boldsymbol{G}$ is shown in Fig.
$\ref{fig2}$.
Details will be discussed in the following subsections.

\subsection{Multi-Label Text Classification}\label{T}
To conduct the TTF task, it is of vital importance to have sufficient facial attribute labels to fully describe a face.
We propose to use the CelebA \cite{b11} dataset which includes 40 facial attribute labels for each face.
To map the free-form natural language descriptions to the 40 facial attributes, we propose to fine-tune a multi-label text classifier $\boldsymbol{T}$ to obtain text embeddings of length 40.
Note that the keywords in the descriptions are the same words or synonyms of text labels in the CelebA dataset.
Some labels might be considered offensive to some people, but we need to use these labels so we can compare our approach to the work of others --- we truly do not wish to cause offence even to synthesized people.

\begin{figure}[!h]
  \centerline{\includegraphics [width=3in]{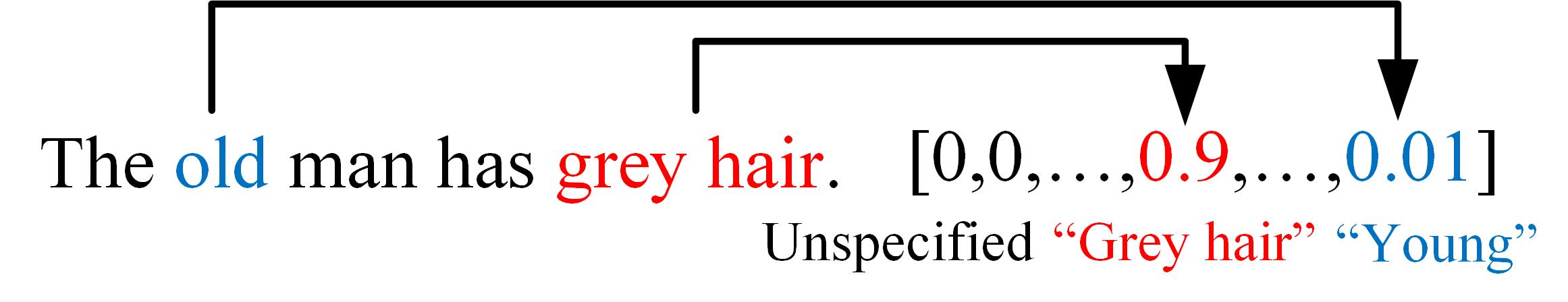}}
  \caption{A possible classification result of the text classifier $\boldsymbol{T}$.}
  \label{sen}
  \vspace{-0.3 cm}
\end{figure}

With these considerations, we adopt the state-of-the-art natural language processing model, Bidirectional Transformer (BERT) \cite{b9}.
In light of the fact that this is a 40-class classification task, we choose to use the large network of the BERT model as it has better performance on high-dimensional training data.
Some features have different names for their opposites attributes.
For example, when training the model $\boldsymbol{T}$,  the feature ``age'' could be represented by either “young" or “old" where, for example, “young" might be represented by a value close to 1 and “old" might be close to 0.
If a feature isn't specified, it is set to 0.
This process is shown in Fig.~\ref{sen}.
Finally, the classifier outputs a text vector of length 40 for each description.

Note that one advantage of our text classifier compared to the earlier text encoders is that there is no restriction on the length of text descriptions.
In previous works, the text is mostly crammed into one or two sentences.
For face descriptions, length is generally much longer than for bird and flower descriptions, which makes traditional text encoders inappropriate.

\subsection{Image Multi-Label Embeddings}
In the proposed framework, an image encoder $\boldsymbol{E}$ is required to predict the feature labels of the generated images.
To do this, we fine-tune a MobileNet model \cite{b10} with the samples of CelebA \cite{b11}.
The reason for choosing MobileNet is that it is a light-weight network model that has a good trade-off between accuracy and speed.
With this model, we can obtain the image embeddings which have the same length as the text vectors of the images generated from the noise vectors.

\subsection{Feature Axes}
After training the image encoder, now we can find the relationship between the noise vectors and the predicted feature labels by logistic regression.
The length of the noise vectors is 512 ($\boldsymbol{x} \in \mathbb{R}^{512}$)  and the length of thefeature vectors is 40 ($\boldsymbol{y} \in \mathbb{R}^{40}$).
Therefore,  we  obtain:

\begin{equation} \boldsymbol{y} = \boldsymbol{x}\cdot\boldsymbol{B}\label{eq1} \end{equation} 

\noindent where $\boldsymbol{B}$ is the matrix of dimension 512$\times$40 to be solved.

This matrix needs to be orthogonalised because we must disentangle all the attributes so that the noise vectors can move along a certain feature axis without affecting the others.
By the Gram-Schmidt process, the projection operator is:

\begin{equation} \mathrm{proj}_\mathbf{u} (\mathbf{v}) = \frac{\left \langle \mathbf{v},\mathbf{u} \right \rangle}{\left \langle \mathbf{v},\mathbf{v} \right \rangle}\mathbf{u}\label{eq2} \end{equation}

\noindent where $\mathbf{v}$ is the axis to be orthogonalised and $\mathbf{u}$ is the reference axis.
Then, we obtain:

\begin{equation} \begin{aligned}  & \mathbf{u}_k = \mathbf{v}_k - \sum_{j=1}^{k-1} \mathrm{proj}_{\mathbf{u}_j}(\mathbf{v}_k), \\  & \mathbf{w}_k = \frac{\mathbf{u}_k}{\left \| \mathbf{u}_k \right \|},\left ( k=1,2,.
    ..40 \right ).
  \end{aligned}
  \label{eq3}
\end{equation}

\noindent In \eqref{eq3}, the matrix $\boldsymbol{W}=[\mathbf{w}_1, \mathbf{w}_2,.
  ..\mathbf{w}_k]$ is normalised so that $\boldsymbol{W}$ is unitary.

After these steps, we obtain the feature axes which are used to guide the update direction of the input noise vectors to obtain the desired features in the output images.

\subsection{Noise Vector Manipulation}

Manipulating the noise vectors is vital to our work because this determines whether the output images will have the described features as the text corpus.
In the model diagram Fig.~\ref{fig2}, this is the process of changing the random noise vector from $\boldsymbol{z}$ to $\boldsymbol{\hat{z}}$ by \eqref{eq4} where $l$ is a column vector which determines the direction and magnitude of the movement along the feature axes.

\begin{equation}
  \boldsymbol{\hat{z}}= \boldsymbol{z} + \boldsymbol{W}\cdot l
  \label{eq4}
\end{equation}

To ensure that the model will produce an image with the desired features no matter where the noise vectors are located in the latent space, we introduce four operations.

\textbf{Differentiation.}
As shown in Fig.~\ref{fig2}, the text classifier embedding output is denoted $l_{trg}$ and the predicted embedding from the initial random vector is given by $l_{org} = \boldsymbol{E}(\boldsymbol{G}(\boldsymbol{z}))$.
Intuitively, we can use $l_{trg}$ to guide the movement of noise vectors along the feature axes.
However, the value range of $l_{trg}$ is $\left [ 0,1 \right ]$.
This means that the model cannot render features in the opposite direction, say, young versus old, because there are no labels corresponding to the opposite values.
To solve this, we use differentiated embeddings $l_{diff}$ to guide the feature editing obtained by \eqref{eq5}

\begin{equation} l_{diff} = l_{trg} - l_{org}.
  \label{eq5}
\end{equation}



In this way, the noise vectors can be moved in both positive and negative directions along the feature axes because the value range of the differentiated embeddings is $\left[ -1,1\right]$.
For the features which have a similar probability value in both the text embeddings and the image embeddings, their probability value is cancelled out and they will not be rendered repeatedly in the output images.
This operation is shown in Fig.~\ref{fig2}.
For each feature, according to its probability level in $l_{trg}$ and $l_{org}$, the movement direction can be positive, negative or neutral.

Note that to minimize interference of the unspecified features in the text descriptions, we do not apply the differentiation operation to such features.
Instead we keep their value as zero in the differentiated embeddings.

\textbf{Nonlinear Reweighting:}
In the differentiated embeddings, the labels with values approaching -1 or 1 are the specified features where the text descriptions are specified in either a positive or negative way.
Apart from these labels, there may be some other labels whose values are between -1 and 1 which tend to interfere with the desired feature rendering.
Therefore, we need to emphasize the specified features.
To do this, we scale the differentiated embeddings range slightly from $\left[ -1,1\right]$ to $\left[-\frac{\pi}{3}, \frac{\pi}{3} \right ]$.
Then we compute the tan(.) of the mapped differentiated embeddings.
As a result, values approaching the ends of the range will get a higher weighting.
In our case, since $\tan(pi/3)= \sqrt{3}$, the reweighed value range is now $\left[ -\sqrt{3}, \sqrt{3} \right ]$.

\begin{figure*}[t!]
  \centerline{\includegraphics [width=\textwidth]{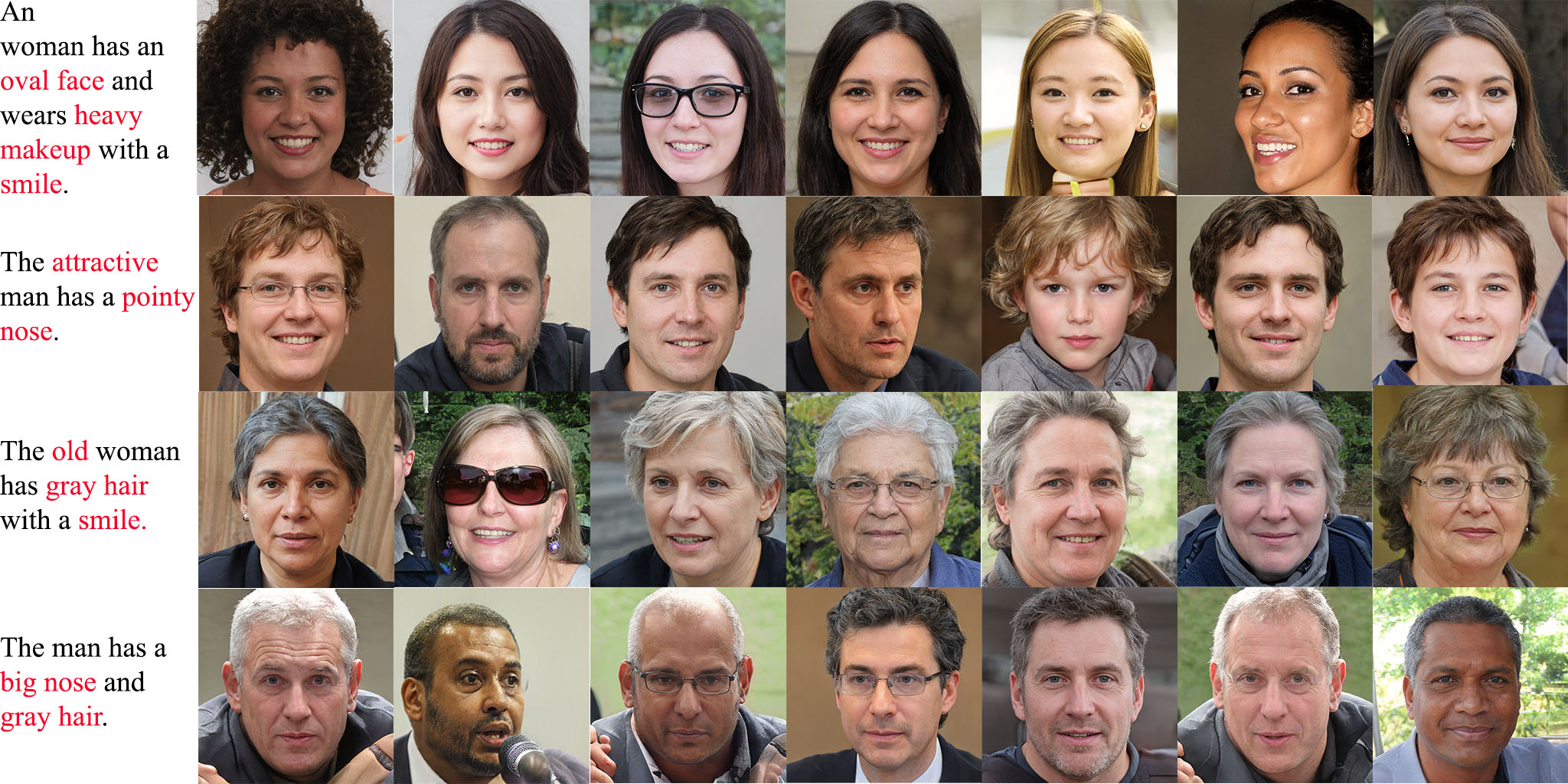}}
  \caption{Images produced with single-sentence input.
    With fewer specified labels in the text, the model generates samples with higher diversity.
  }
  \label{unt1}
\end{figure*}

\textbf{Normalization:}
As the noise vectors are sampled from a normal distribution, they have a higher probability to be sampled near the origin where the probability density is high.
However, the more steps we move the vectors along different feature axes, the larger the distance becomes between these vectors and the origin, which will lead to more artifacts in the generated images.
That is why we need to renormalise the vectors after each movement along the axes.
This distance can be denoted as  $L_1$ distance.
Therefore, for the noise vector $\boldsymbol{X} = [\mathbf{x}_1, \mathbf{x}_2,.
  ..\mathbf{x}_n]$, we get $\boldsymbol{X}^\prime = [\mathbf{x}_1^\prime, \mathbf{x}_2^\prime,...,\mathbf{x}_n^\prime]$ with \eqref{eq6}
\vspace{-0.1cm}
\begin{equation}
  \begin{aligned}
     & \left \| \mathbf{x} \right \|_1 = \sum_{i=1}^{N=512} \left| \mathbf{x}_i \right|              \\
     & \mathbf{x}_i^\prime = \frac {\mathbf{x}_i} {\left \| \mathbf{x} \right \|_1} (i= 1,2,...,512) \\ \label{eq6}
  \end{aligned}
  \vspace{-0.6cm}
\end{equation}

\textbf{Feature lock:}
To make the face morphing process more stable, we have a feature lock step each time we move the vectors along a certain axis.
In other words, the model only uses the axes along which the vectors have been moved as the basis axes to disentangle the following feature axis.
While for other axes of unspecified attributes in the textual descriptions, the movement direction and step size along such axes are not fixed to ensure diversity in the generated images.
In this way, noise vectors are locked only in terms of the features mentioned in the descriptions.

\subsection{High Resolution Generator}\label{HRG}
The generator $\boldsymbol{G}$ we use is a pre-trained model of StyleGAN2 \cite{b12}.
On the basis of mapping the noise vectors which are sampled from the normal distribution to the intermediate latent space, StyleGAN2 improves the small artifacts by revisiting the structure of the network.
With this generator, not only can the model synthesise high-resolution images, but it can also render the desired features from the manipulated input vectors.

\begin{figure*}[t!]
  \centerline{\includegraphics [width=\textwidth]{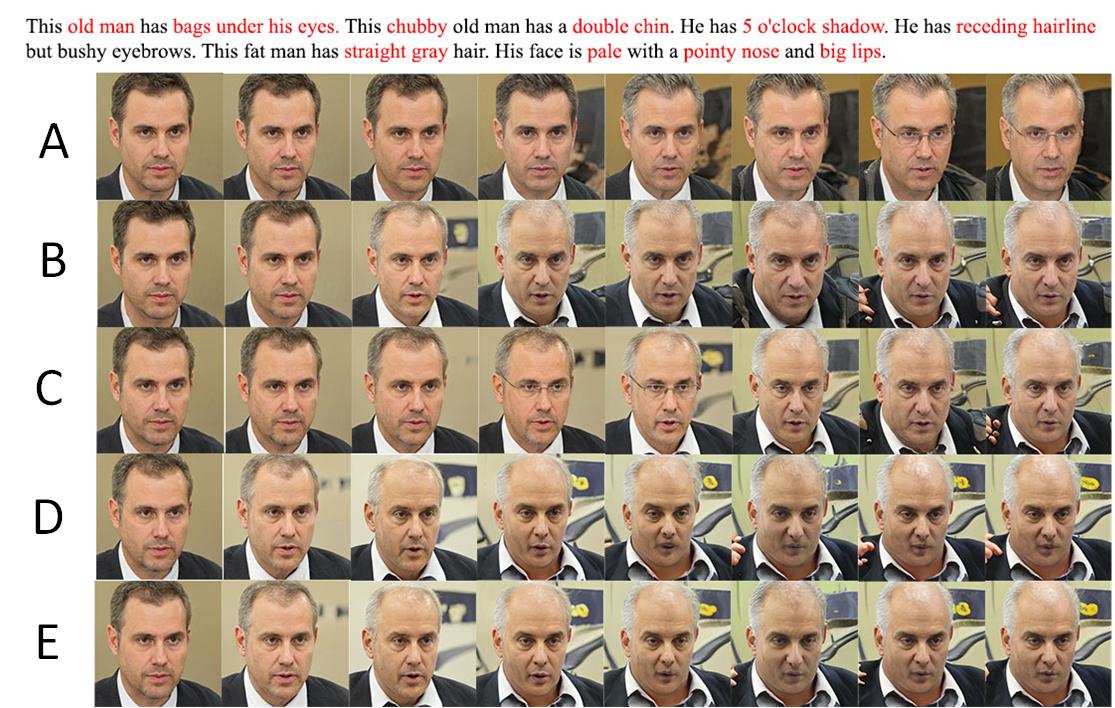}}
  \caption{Image morphing process of each group in ablation study. (A) A group with all operations. (The default setting for TTF-HD) (B) A group with reweighting, differentiation, and normalisation operations. (C) A group with reweighting, differentiation operations. (D) A group with the reweighting operation. (E) Blank group.
    We fix the noise vector input of each group.
    The figure shows the morphing process from the random image on the left column to the final output on the right column.
  }
  \label{fig5}
\end{figure*}

\section{Experiments $\&$ Evaluation}
\textbf{Dataset:}
The dataset we use is CelebA \cite{b11} which contains over 200k face images.
For each sample, there is a paired one-shot label vector whose length is 40.
In addition, there is another paired text description corpus set in which every description has up to 10 sentences.
There may be some redundant sentences in some of them, but every description includes all the features the paired label vector indicates.
We use this dataset to fine-tune the pre-trained multi-label text classifier and the pre-trained image encoder.

\textbf{Experimental setting:} In our evaluation experiments, we randomly choose 100 text descriptions.
With each of them, the model will randomly generate 10 images.
Therefore, the test set has 1000 images in total.
As the experiments show, there will be significant image morphing when the noise vector moves twice along certain feature disentangled axis.
Thus, we set the step size as 1.2, which multiplies the reweighted output of the differentiated vector.
This guarantees a final weight which is used to move along the axis of around 2 ($\sqrt{3}\times1.2$).

\subsection{Qualitative Evaluation}

\textbf{Image quality:} Fig.
\ref{fig1} also shows the paired descriptions in each group.
We can see that most of the generated images are correctly rendered with the specified features.

\textbf{Image diversity}.
To show the proposed method has great feature generalisation capacity, we conduct the image synthesis conditioned on the single-sentence description.
In other words, apart from the key features that the sentences describe, the model should diversify the other facial features in the output.
As Fig.~\ref{unt1} shows, for each single-sentence description, the proposed model produces images with high diversity.

\subsection{Quantitative Evaluation}
In this section, we apply three metrics to evaluate the above three criteria respectively.
They are Inception Score (IS) \cite{b25} which is used in many previous works, Learned Perceptual Image Patch Similarity (LPIPS) \cite{b22} which is for evaluating the diversity of the generated images, and Cosine Similarity which is widely used to evaluate the similarity of two chunks of a corpus in natural language processing.
Due to the lack of the source code for most of the works in the TTF area such as T2F 2.0 \cite{b17}, we compare our experimental results with the TTF implementation of AttnGAN \cite{b8} which has produced the best results so far.

\begin{table}[htb]
  \caption{Evaluation results of different models}
  \begin{center}
    \begin{tabular}{l|ccccc}
      \hline
      \cline{2-4}
      \textbf{Methods} & \textbf{\textit{IS}}     & \textbf{\textit{CS}*} & \textbf{\textit{LPIPS}} \\
      \hline
      TTF-HD (ours)    & \textbf{1.117$\pm$0.127} & \textbf{0.664}        & 0.583$\pm$0.002         \\
      AttnGAN          & 1.062$\pm$0.051          & 0.511                 & ------                  \\
      \hline
      \multicolumn{4}{l}{*Maximum for each group.}
    \end{tabular}
    \label{tab2}
  \end{center}
\end{table}

Table~\ref{tab2} shows the evaluation results of different models.
We see the proposed TTF-HD method outperforms the state-of-the-art method AttnGAN \cite{b8} in terms of both image quality and Text-to-Image similarity.

\subsection{Ablation Study}
In Section 3, we propose four operations to manipulate the noise vector to get the desired features.
In this subsection, we conduct the ablation study and discuss the effects of the different operations applied.

To conduct the ablation study, we have 5 experiment settings.
We choose one face description and produce 100 random images under each experimental setting respectively.
Then, we use the above three metrics to evaluate the effect of different operations.

Fig.
\ref{fig5} shows the morphing process of the generated images.
We can see that with the proposed four manipulating operations, Group A can finally obtain an output with all desired features.
While for other groups, the final morphing images all suffer from artifacts on the rendering of the face and the background.
This is because with too many feature axis moving steps, the noise vector has been moved to a low-density region of the latent space distribution, which leads to the mode collapse problem.
\begin{table}[htb]
  \caption{Ablation study evaluation results}
  \begin{center}
    \begin{tabular}{l|ccccc}
      \hline
      \textbf{Exp.}     & \multicolumn{3}{|c}{\textbf{Evaluation Metrics}}                                                    \\
      \cline{2-4}
      \textbf{Settings} & \textbf{\textit{IS}}                             & \textbf{\textit{CS}*} & \textbf{\textit{LPIPS}}  \\
      \hline
      Group A           & 1.122$\pm$0.043                                  & 0.754                 & \textbf{0.634$\pm$0.005} \\
      Group B           & 1.116$\pm$0.080                                  & 0.739                 & 0.608$\pm$0.005          \\
      Group C           & \textbf{1.187$\pm$0.062}                         & \textbf{0.762}        & 0.603$\pm$0.005          \\
      Group D           & 1.101$\pm$0.095                                  & 0.683                 & 0.521$\pm$0.006          \\
      Group E           & 1.102$\pm$0.033                                  & 0.706                 & 0.532$\pm$0.005          \\
      \hline
      \multicolumn{4}{l}{*Maximum for each group}
    \end{tabular}
    \label{tab3}
  \end{center}
\end{table}

Table~\ref{tab3} shows the quantitative evaluation metrics on different groups of TTF-HD.
We can see that Group A has the best diversity score as well as the second-best performance in terms of both IS and CS score.
This suggests that applying all proposed operations leads to a good trade-off between image quality, text-to-face similarity and diversity.

\section{Conclusion}

In this paper, we set three main goals in the text-to-face image synthesis task: 1) High image resolution, 2) Good text-to-image consistency, and 3) High image diversity.
To this end, we propose a model, named TTF-HD, comprising a multi-label text classifier, an image encoder, a high-resolution image generator, and feature-disentangled axes.
From both qualitative and quantitative evaluative comparisons, we  see the generated images exhibit good image quality, text-to-image similarity, and image diversity.

However, the model is still not entirely robust.
There are always some images in the batch that are far more consistent with the text descriptions.
This is possibly caused by insufficient accuracy of the text classifier and image encoder due simply to the lack of training data.
In addition, features in the latent space are still not well disentangled, so that when you are moving the noise vector along one feature axis, other features which are highly correlated with it may also change.
These issues must be addressed in future research.

\end{document}